\documentclass[9pt, conference]{IEEEtran}
\IEEEoverridecommandlockouts

\usepackage{cite}
\usepackage{amsmath,amssymb,amsfonts}
\usepackage{algorithmic}
\usepackage{graphicx}
\usepackage{textcomp}
\usepackage{xcolor}

\usepackage{booktabs}
\usepackage{multirow}
\usepackage{multicol}
\usepackage{cleveref}
\usepackage{amsmath,amsfonts,bm}

\graphicspath{{./images}}
\def\BibTeX{{\rm B\kern-.05em{\sc i\kern-.025em b}\kern-.08em
    T\kern-.1667em\lower.7ex\hbox{E}\kern-.125emX}}

\begin{document}

\title{Learning Semantic Facial Descriptors for Accurate Face Animation
\thanks{\IEEEauthorrefmark{2}Corresponding Author}
}


\author{\IEEEauthorblockN{Lei Zhu$^1$, Yuanqi Chen$^1$, Xiaohang Liu$^1$, Thomas H.Li$^1$, Ge Li$^1$\IEEEauthorrefmark{2}}
\IEEEauthorblockA{
\textit{$^1$School of Electronic and Computer Engineering, Shenzhen Graduate School, Peking University}\\
}
}

\maketitle

\begin{abstract}
Face animation is a challenging task. Existing model-based methods (utilizing 3DMMs or landmarks) often result in a model-like reconstruction effect, which doesn't effectively preserve identity. Conversely, model-free approaches face challenges in attaining a decoupled and semantically rich feature space, thereby making accurate motion transfer difficult to achieve. We introduce the semantic facial descriptors in learnable disentangled vector space to address the dilemma. The approach involves decoupling the facial space into identity and motion subspaces while endowing each of them with semantics by learning complete orthogonal basis vectors. We obtain basis vector coefficients by employing an encoder on the source and driving faces, leading to effective facial descriptors in the identity and motion subspaces. Ultimately, these descriptors can be recombined as latent codes to animate faces. Our approach successfully addresses the issue of model-based methods' limitations in high-fidelity identity and the challenges faced by model-free methods in accurate motion transfer. Extensive experiments are conducted on three challenging benchmarks (i.e. VoxCeleb, HDTF, CelebV). Comprehensive quantitative and qualitative results demonstrate that our model outperforms SOTA methods with superior identity preservation and motion transfer.
\end{abstract}

\section{Introduction}\label{sec:intro}




Due to the remarkable progress of deep generative models, face animation has drawn increased attention, where the goal is to transfer the motion~(3D head orientation and expression) of a target face to a source. 
Such capabilities give rise to numerous applications, including virtual reality and film production. 
However, it's challenging to achieve accuracy in both aspects simultaneously: accurately preserving source identity and transferring target motion.

Many model-based methods \cite{meshguided,dual,faceformer,2022face2face,2023audio,yu2023nofa,PIRendererCP,jin2023diffusionret,HifiFace3S,face2faceρ,StyleHEATOH,Hifihead, metaportrait,DGFR} leverage
3DMMs \cite{3DMM,3dmm2} or landmark detectors \cite{3dlandmark,metaportrait} to obtain disentangled facial parameters or landmarks containing motion information.
3DMM \cite{3DMM} is built from a large corpus of 3D face scans, using PCA bases to derive a parametric model of facial shape and texture variation. 
However, the coarse face shapes generated through the computer graphics techniques used in 3DMMs have limited ability to represent the full range of real face shape variations, leading to model-like and not highly realistic results.
A limitation of landmark-based methods \cite{metaportrait,DGFR,liu2021li} is that landmarks preserve identity, thus impeding their applicability to cross-id face animation.
In addition, using pretrained models can easily lead to error accumulation, especially when the training data for face animation differs from the distribution of pretrained models. Furthermore, using pretrained models suffers from a tedious big-budget construction procedure.

Model-free methods \cite{li2023weakly,FOMM,motionrep,vid2vid} learn to reconstruct driving frames by warping source images utilizing predicted dense flow fields alongside learned explicit facial structural features like landmarks \cite{FOMM,vid2vid} or regions \cite{motionrep}.
Nevertheless, these methods struggle to achieve accurate motion transfer and generate numerous artifacts.
The reason is that they have difficulty obtaining a highly disentangled feature space that is semantically meaningful. They also have difficulty establishing high-resolution flow fields, limiting applicability to high-resolution animation. Despite \cite{hong2023implicit} employing attention mechanisms for texture completion, such methodologies still encounter challenges in maintaining the fidelity of the identities in the source image.

\begin{figure}[t]
\centering
  \includegraphics[width=1.0\linewidth]{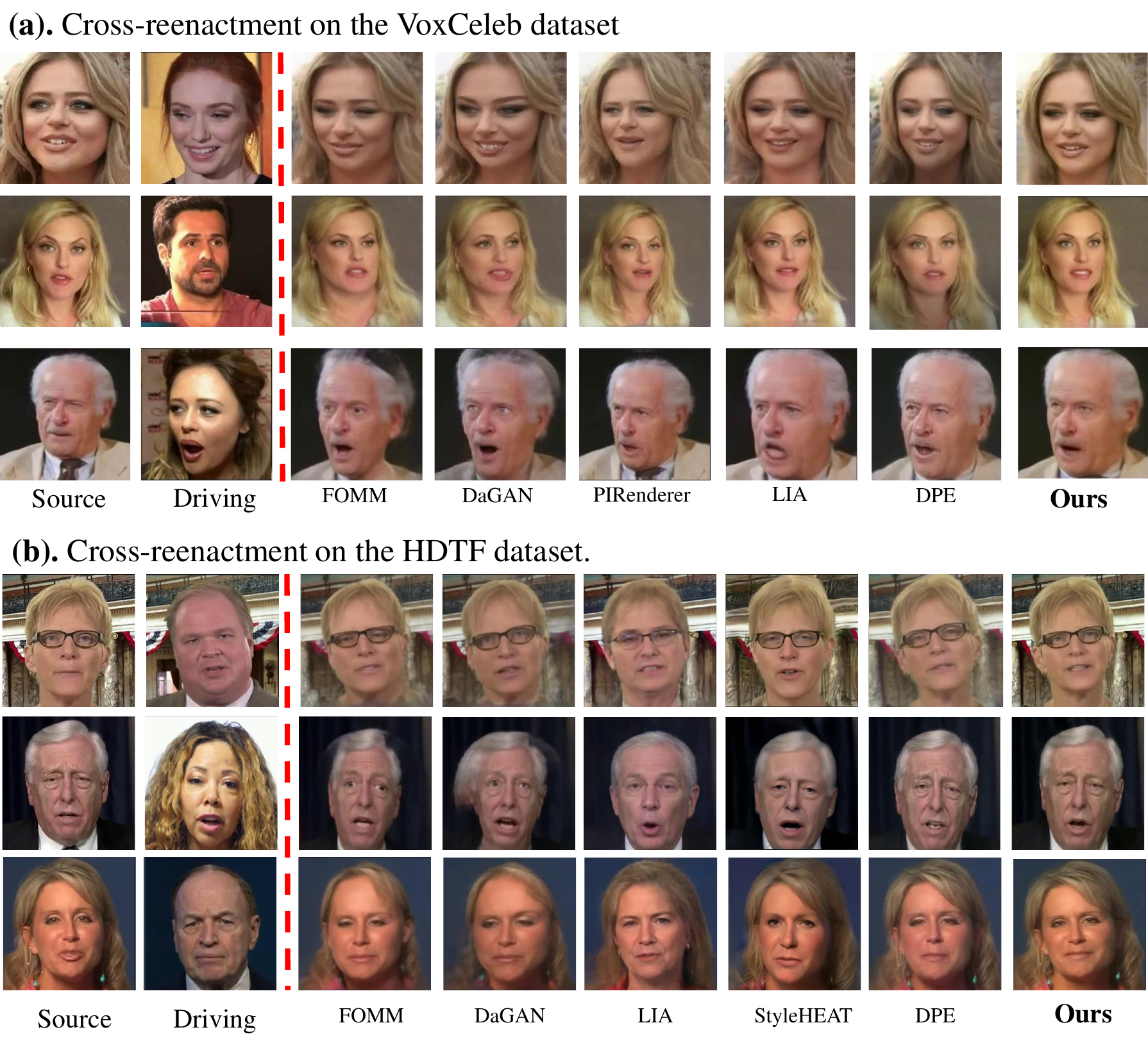}
  \vspace{-0.7cm}
  \caption{Qualitative comparisons with SOTA methods on Cross-reenactment.}
  \label{Qualitative comparisons with SOTA methods on Cross-reenactment.} 
  \vspace{-0.6cm}
\end{figure}

Unlike these model-based or model-free methods that use explicit structural information, we obtain a face representation that is highly disentangled and semantically meaningful in identity and motion. 
Our approach effectively addresses the issue of model-based methods' limitations in achieving high fidelity, and the challenge faced by model-free methods in accurately realizing motion transfer.
Inspired by 3DMM's use of PCA bases to build a parametric model of facial shapes, we wondered: \textbf{\textit{Can we learn a complete set of orthogonal basis vectors to construct high-level disentangled and semantically meaningful face representations with high fidelity?}}

\begin{figure*}[ht] 
    \centering 
    \includegraphics[scale=0.43]{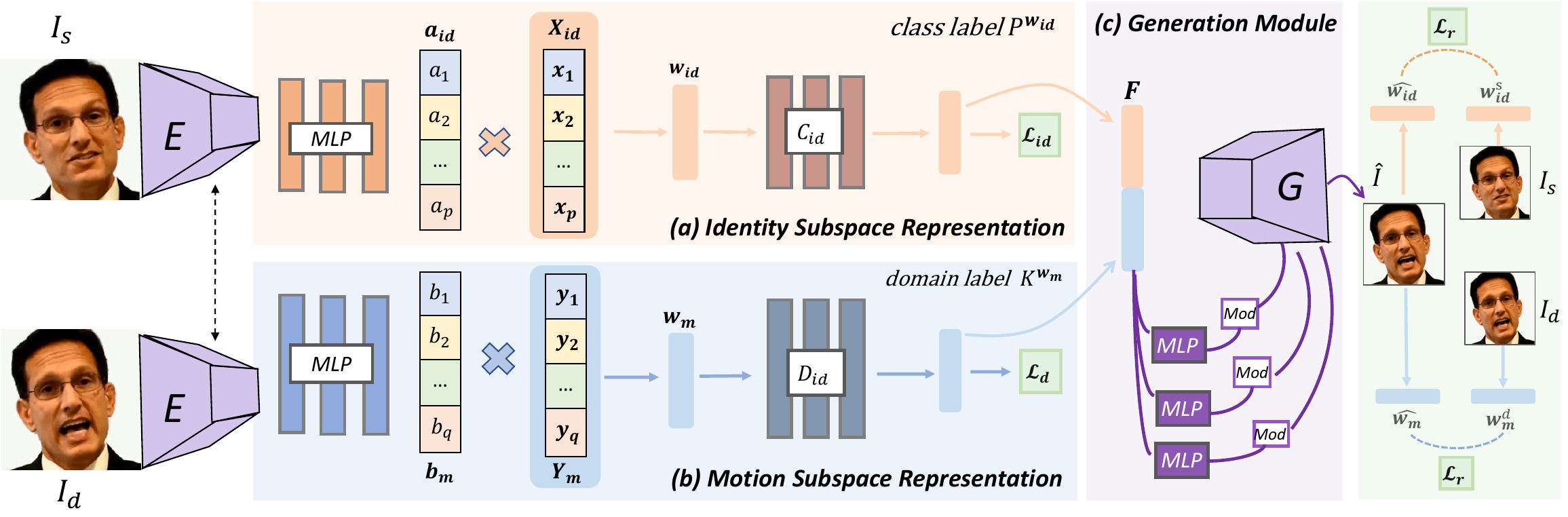} 
    \vspace{-0.3cm}
    \caption{The pipeline of our method. We take source image $I_s$ and driving image $I_d$ as input and output animated image $\hat{I}$. Refer to \ref{sec:method} for the specific process.} 
    \vspace{-0.5cm}
    \label{Overview of our proposed model} 
\end{figure*}

Based on this problem, we learn a highly disentangled vector space in a self-supervised manner and build high-level semantic face descriptors for face animation. According to the requirement of disentangling identity and motion, we assume the face feature space is composed of an identity subspace and a motion subspace, each spanned by a set of complete orthogonal basis vectors. We devise a series of learning strategies to update the vectors so that the two subspaces are decoupled from each other while retaining semantic. We can subsequently perform distinct linear combinations on these basis vectors within the subspaces. The weights of basis vectors encoding specific identities and motions are obtained by effectively extracting information from the source and driving images using encoders, which are optimized during network training. Finally, we combine the identity and motion descriptors from the two subspaces as latent codes to control the generation process. We can obtain a series of intermediate states between two motions by simply adjusting the coefficients.

We train from scratch on datasets of two different resolutions (i.e. VoxCeleb, HDTF) and achieve high-fidelity animation results on three test datasets with precise motion transfer. Experiments demonstrate our results far surpass other SOTA methods on wild-face images. The characteristics of our method are summarized as follows:

\begin{itemize}
  \setlength{\itemsep}{0.3pt}
  \setlength{\parsep}{0.3pt}
  \setlength{\parskip}{0.3pt}
  \item High-fidelity: Due to independence from patterned structural information and highly decoupled identity descriptors, we can achieve superior fidelity in identity compared to SOTA model-based approaches.
  \item High-precise: The motion descriptors possess abundant semantic information, enabling us to achieve more accurate motion transfer compared to SOTA model-free approaches.
  \item High-efficiency: The space we have constructed is linear, enabling straightforward linear interpolation between two motions. We require no reliance on pretrained models, leading to efficient inference. Additionally, the space we have constructed can be successfully extended to high-resolution image animations.
  \vspace{-0.1cm}
\end{itemize}

\section{Method}\label{sec:method}
\subsection{Face Representation Space}
The framework of our proposed method is shown in Fig.\ref{Overview of our proposed model}.
Instead of using an explicit 3D graphics model, we directly represent human faces in a high-dimensional vector space for 2D face animation. 
We treat human faces as signals in the physical world. 
Based on the theory of signal processing, all signals in a vector space can be represented by a complete set of orthogonal basis vectors.
Inspired by this characteristic, our idea is to learn a set of orthogonal basis vectors $\mathbf{D}=\left\{\mathbf{x}_1, \ldots, \mathbf{x}_{\mathbf{p}},\mathbf{y}_{\mathbf{1}},\ldots,  \mathbf{y}_{\mathbf{q}} \right\}$, where a portion of the basis vectors (\textbf{identity subspace}) represents identity
$\mathbf{X_{id}}=\left\{\mathbf{x}_1, \ldots, \mathbf{x}_{\mathbf{p}}\right\},$
while the remaining ones capture facial expressions and head poses (\textbf{motion subspace})
$\mathbf{Y_{m}}=\left\{\mathbf{y}_{1}, \ldots, \mathbf{y}_{\mathbf{q}}\right\}.$
any two basis vectors $\mathbf{d}_{\mathbf{i}},\mathbf{d}_{\mathbf{j}} \in \mathbf{D}$ follow the constrain

\begin{equation}
    \langle \mathbf{d}_{\mathbf{i}}, \mathbf{d}_{\mathbf{j}} \rangle = 0 \, \mathbf{if}\: i \neq j; \; 1 \,\mathbf{if}\: i = j
\end{equation}

In our method, we treat $\mathbf{D}$ as a learnable matrix and incorporate the Gram-Schmidt process during each forward pass, ensuring the fulfillment of the orthogonality constraint.
We then combine each vector in identity basis with a vector $\mathbf{a_{id}}=\left\{a_{1}, \ldots, a_{p}\right\}$ and in motion basis with a vector $\mathbf{b_m}=\left\{b_{1}, \ldots,{b}_{q}\right\}$.
Therefore, the face can be modeled as
\begin{equation} 
\mathbf{F}= \mathbf{a_{id}}\mathbf{X_{id}} + \mathbf{b_m}\mathbf{Y_{m}} =\sum_{i=1}^p a_i \mathbf{x}_{\mathbf{i}}+\sum_{i=1}^{q} b_i \mathbf{y}_{\mathbf{i}}
\end{equation}

\begin{figure}[ht] 
    \centering 
    \includegraphics[width=0.85\linewidth]{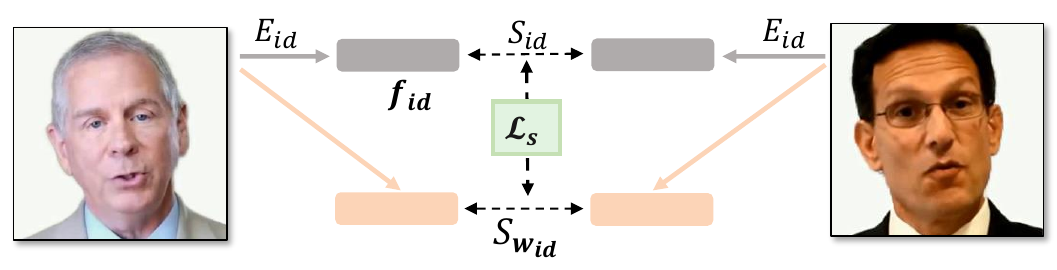} 
    \vspace{-0.3cm}
    \caption{The pipeline of our identity subspace distillation.} 
    \label{The pipeline of our identity subspace distillation.} 
    \vspace{-0.2cm}
\end{figure}

where $\mathbf{x}_{\mathbf{i}} \in \mathbb{R}^N$ and $a_i \in \mathbb{R}$ for all $i \in\{1, \ldots, p\}$,$\mathbf{y}_{\mathbf{i}} \in \mathbb{R}^N$ and $b_i \in \mathbb{R}$ for all $i \in\{1, \ldots, q\}$. $a_i,b_i$ denotes the weight coefficients of basis vectors corresponding to identity subspace and motion subspace respectively. 
Supposing that $\mathbf{w_{id}}=\mathbf{a_{id}} \mathbf{X_{id}}$ (\textbf{identity subspace descriptor}) corresponds to the identity latent code and $\mathbf{w_{m}}=\mathbf{b_{m}} \mathbf{Y_{m}}$ (\textbf{motion subspace descriptor}) corresponds to the motion latent code in the latent space of generation network.

\subsection{Disentangling Identity and Motion Subspaces} 
To attain complete disentanglement of the identity and motion subspaces, we adopt two mutually reinforcing learning strategies. Firstly, by capitalizing on the feature extraction capabilities of a pretrained facial recognition network, we align the distribution of identity subspace representations with that of true identity features. This allows us to effectively distill identity information in face space. Secondly, we incorporate domain adversarial techniques to eliminate identity-related information from the motion subspaces derived from the driving images. 



\subsubsection{Identity Subspace Distillation.} 
We enhance the identity similarity between images with different motions but the same identity, aiming to incorporate as much identity-related information as possible into the identity subspace representation as shown in Fig.\ref{The pipeline of our identity subspace distillation.}.
By adopting a self-supervised learning strategy, we divide a batch of $B$ samples $\left\{I_s^1, \ldots, I_s^{B}\right\}$ into $T$=$\frac{B}{2}$ pairs $\left\{  (I_s^1, I_s^{T+1} ), \ldots,  (I_s^T, I_s^B) \right\}$. 
From this, we obtain $T$ sets of identity subspace representations $\left\{  (\mathbf{w}_{id}^1, \mathbf{w}_{id}^{T+1} ), \ldots,  (\mathbf{w}_{id}^T, \mathbf{w}_{id}^B) \right\}$ and compute the similarity for each group, resulting in an identity subspace representation similarity vector $\mathbf{s}_{\mathbf{w}_{id}} = (  s_{\mathbf{w}_{id}}^1, \ldots, s_{\mathbf{w}_{id}}^{T} )$ of length $T$. 
For each binary tuple, we extract identity features using the pre-trained face recognition network ArcFace \cite{ArcFace} $E_{id}$, resulting in $T$ sets of identity features$\left\{  (\mathbf{f}_{id}^1, \mathbf{f}_{id}^{T+1} ), \ldots,  (\mathbf{f}_{id}^T, \mathbf{f}_{id}^B) \right\}$. We then compute the similarity for each group, yielding a target identity similarity vector $\mathbf{s}_{id} = ( s_{id}^1, \ldots, s_{id}^T)$ of length $T$. 

The features extracted by the face recognition network provide us with discriminative identity information. By improving the cosine similarity between the identity subspace representation similarity vector $\mathbf{s}_{\mathbf{w}_{id}}$ and the target identity similarity vector $\mathbf{s}_{id}$ mentioned above, the identity subspace representation can be made with more discriminative and higher-level identity information. We define the identity similarity loss $\mathcal{L}_{s}$ as follows.

\begin{equation}
\mathcal{L}_{s} =-\frac{\mathbf{s}_{id} \cdot \mathbf{s}_{\mathbf{w}_{id}}}{\max \left(\left\|\mathbf{s}_{id}\right\|_2 \cdot\left\|\mathbf{s}_{\mathbf{w}_{id}}\right\|_2, \epsilon\right)}
\end{equation}
Where $\epsilon$ is a small constant. 
Finally, we maximize the cosine similarity between the aforementioned two identity similarity vectors.

\subsubsection{Eliminating identity in motion subspace.}
In order to ensure that the motion subspace representation does not contain the driving's identity information, we use an identity eraser to eliminate any identity information present in the motion subspace. 
Specifically, we design a domain discriminator with a 3-layer MLP as the identity eraser. It takes the motion subspace representation $\mathbf{w_{m}}$ as input and infers its identity domain label as:
\begin{equation}
K^{\mathbf{w_{m}}}=D_{id}(\mathbf{w}_{m})
\end{equation}
Where $D(\cdot)$ denotes the MLP, $K^{\mathbf{w_{m}}} \in \mathbb{R}^C$ denotes the predicted identity label, $C$ denotes the number of identities,we optimize the MLP with cross-entropy loss as:
\begin{equation}
\mathcal{L}_d=H(K, K^\mathbf{w_{m}})
\end{equation}
Where $H(\cdot)$ denotes cross-entropy loss, and $K$ denotes the one-hot coding of driving's identity label, which is assumed as a vector with the length of identity numbers in the training set. Especially, we use a negative hyperparameter $-\lambda_d$ to introduce the domain loss $\mathcal{L}_d$ as an adversarial loss $\mathcal{L}_{adv}$ into the motion subspace learning to erase any driving's identity. This loss is then used to iteratively optimize both the motion representation generator and the domain discriminator. When the domain discriminator has strong discriminative capabilities, if the loss persists, it can be inferred that identity information in the motion representation has been erased.


\subsection{Enriching Semantic Information in two Subspaces} 
After separating the two spaces, in order to sufficiently express the semantic in both the identity and motion subspaces, we apply latent regression constraints to the identity subspace and motion subspace, as well as an additional category constraint specifically to enhance identity authentication.


\subsubsection{Latent Regression Constraint.}
In order to ensure that the identity of the final generated image $\hat{I}$ remains consistent with the source image $I_s$ and the motion aligns with the driving image $I_d$, we employ the following latent regression losses $\mathcal{L}_{r}$ to assist in optimizing two subspaces. This facilitates the sufficient expression of semantic information in both the identity and motion subspaces.
\begin{equation}
\mathcal{L}_{r}=\left\| \hat{\mathbf{w}}_{id}- \mathbf{w}_{id}^s \right\|_1 + \left\| \hat{\mathbf{w}}_{m}- \mathbf{w}_{m}^d \right\|_1
\end{equation}
Where $\hat{\mathbf{w}}_{id}$ and $\mathbf{w}_{id}^s$ respectively denote the identity descriptor extracted from the final generated image $\hat{I}$ and the source image $I_s$, $\hat{\mathbf{w}}_{m}$ and $\mathbf{w}_{m}^d$ respectively denote the motion descriptor extracted from the final generated image $\hat{I}$ and the driving image $I_d$.
.
\subsubsection{Identity Authentication Enhancement.}
We can further optimize the identity subspace representation $\mathbf{w}_{id}$ through an identity classifier $C_{id}$ using the softmax cross-entropy loss $\mathcal{L}_{id}$ to enhance identity authentication. Specifically, the identity classifier produces the probability $P^{\mathbf{w}_{id}} \in \mathbb{R}^C$ that the identity representation belongs to each identity, $C$ denotes the number of identities.
\begin{equation}
\mathcal{L}_{id}=-\log \left(P_i^{\mathbf{w}_{id}}\right)
\end{equation}
where $i$ is the category of the ground-truth identity label.

Once we obtain the face representation $\mathbf{F}=\mathbf{w}_{id}+ \mathbf{w}_m$, we take it as our input and use $G$ to decode a flow field $\phi_{s \rightarrow d}$ to warp the source feature $x_s^{e n c}=\left\{x_i^{e n c}\right\}_1^N$ extracted in the encoder $E$.

\subsection{Training} 
We train our network in a self-supervised manner to generate $\hat{I}$ using seven losses, i.e., a reconstruction loss $\mathcal{L}_{\text {recon }}$, a perceptual loss $\mathcal{L}_{v g g}$, an adversarial loss $\mathcal{L}_{a d v}$ and four losses mentioned above, i.e., $\mathcal{L}_s$,  $\mathcal{L}_d$,  $\mathcal{L}_{r}$,  $\mathcal{L}_{id}$. Our total loss is given by a combination of the aforementioned loss terms:

\begin{align}
&\begin{aligned}
\mathcal{L}=\lambda_{\text {recon }} \mathcal{L}_{\text {recon }}+ & \lambda_{\text {vgg }} \mathcal{L}_{\text {vgg}}+\lambda_{\text {adv}} \mathcal{L}_{\text {adv}}+ \lambda_{\text {s}} \mathcal{L}_{\text {s}}-\lambda_{\text {d}} \mathcal{L}_{\text {d}} \\
& + \lambda_{\text {r}}\mathcal{L}_{\text {r}}+ \lambda_{\text {id}}\mathcal{L}_{\text {id}}
\end{aligned}
\end{align}
where we use $\lambda_{\text {recon }}=1, \lambda_{\text {vgg}}=1, \lambda_{\text {adv}}=1, \lambda_{\text{s}}=2, \lambda_{\text {d}}=0.04,  \lambda_{\text {r}}=1,  \lambda_{\text {id}}=0.05$.

\begin{table}[t]
 \begin{center}{ \footnotesize
 \centering
 \centering
 \caption{Performance on VoxCeleb and CelebV datasets.}
  \vspace{-0.3cm}
 \scalebox{0.6}[0.6]{
   \begin{tabular}{c|ccccc|ccccc}
      \toprule & CSIM $\uparrow$ &FID $\downarrow$  &SSIM $\uparrow$  & AED $\downarrow$ & APD $\downarrow$  & CSIM $\uparrow$ &FID $\downarrow$  &SSIM $\uparrow$  & AED $\downarrow$ & APD $\downarrow$ \\
      \midrule  & \multicolumn{5}{c}{ VoxCeleb } & \multicolumn{5}{c}{ CelebV} \\
      \midrule   FOMM & $0.480$ & $28.17$ & $0.945$ & $0.149$ & $0.070$ & $0.491$ & $42.30$ & $0.932$ & $0.297$ & $0.105$ \\
      PIRenderer & $0.482$ & $32.60$ & $0.941$ & $0.106$ & $0.069$ & $\textbf{0.506}$ & $41.09$ & $0.940$ & $0.281$ & $0.099$ \\
      DaGAN & $0.466$ & $35.16$ & $0.943$ & $0.135$ & $0.064$ & $0.485$ & $49.99$ & $0.938$ & $0.292$ & $0.093$ \\
      LIA & $0.473$ & $33.58$ & $0.957$  & $0.127$ & $0.066$ & $0.449$ & $52.00$ & $0.941$ & $0.269$ & $0.096$ \\
      DPE & $0.481$ & $31.62$ & $0.957$  & $0.052$ & $0.061$ & $0.476$ & $46.21$ & $0.942$ & $0.266$ & $0.097$ \\
      Ours & $\textbf{0.503}$ & $\textbf{27.81}$ & $0.957$ & $\textbf{0.035}$ & $\textbf{0.054}$ & $0.497$ & $\textbf{39.46}$ & $\textbf{0.945}$ & $\textbf{0.264}$ & $\textbf{0.092}$ \\
       \hline
   \end{tabular}}
   \label{Quantitative comparisons on the VoxCeleb dataset and CelebV dataset.}
   }
    \end{center}
    \vspace{-0.4cm}
\end{table}

\begin{table}[t]
 \begin{center}{ \footnotesize
 \centering
 \centering
 \caption{Quantitative comparisons on the HDTF dataset.}
 \vspace{-0.3cm}
 \scalebox{0.6}[0.6]{
   \begin{tabular}{c|ccccc|ccccc}
      \toprule
      & CSIM $\uparrow$ &FID $\downarrow$  &SSIM $\uparrow$  & AED $\downarrow$ & APD $\downarrow$  & CSIM $\uparrow$ &FID $\downarrow$  &SSIM $\uparrow$  & AED $\downarrow$ & APD $\downarrow$\\
      \midrule & \multicolumn{5}{c}{ Self-reenactment } & \multicolumn{5}{c}{ Cross-reenactment } \\
      \midrule   FOMM & $0.712$ & $\textbf{17.27}$ & $0.935$ & $0.113$ & $0.024$ & $0.621$ & $67.74$ & $0.919$ & $0.311$ & $0.049$ \\
      DaGAN & $0.726$ & $24.53$ & $0.917$ & $0.109$ & $0.019$ & $0.594$ & $69.27$ & $0.906$ & $0.302$ & $0.045$ \\
      LIA & $0.320$ & $53.24$ & $0.812$  & $0.271$ & $0.042$ & $0.229$ & $107.34$ & $0.716$ & $0.408$ & $0.051$ \\
      DPE & $0.754$ & $19.87$ & $0.957$  & $\textbf{0.103}$ & $0.019$ & $0.638$ & $54.75$ & $0.938$  & $\textbf{0.243}$ & $0.035$ \\
      StyleHEAT & $0.631$ & $21.61$ & $0.924$ & $0.142$ & $0.031$ & $0.476$ & $62.95$ & $0.915$ & $0.295$ & $0.037$ \\
      Ours & $\textbf{0.838}$ & $18.80$ & $\textbf{0.969}$ & $\textbf{0.105}$ & $\textbf{0.016}$ & $\textbf{0.724}$ & $\textbf{50.47}$ & $\textbf{0.959}$ & $0.269$ & $\textbf{0.034}$ \\
       \hline
   \end{tabular}}
   \label{Quantitative comparisons on the HDTF dataset}
   }
    \end{center}
    \vspace{-0.4cm}
\end{table}

\begin{table}[t]
\vspace{-0.5cm}
 \begin{center}{ \footnotesize
 \centering
 \caption{Ablation results of cross-reenactment on VoxCeleb.}
 \vspace{-0.2cm}
 \scalebox{0.9}[0.9]{
   \begin{tabular}{c|ccccc}
      \toprule & CSIM $\uparrow$ & FID $\downarrow$  & SSIM $\uparrow$ & AED $\downarrow$ & APD $\downarrow$ \\
      \midrule  Baseline & $0.473$ & $33.58$ &$0.957$ & $0.127$ & $0.066$ \\
      +Subs. & $0.483$ & $32.10$ & $0.957$ & $0.072$ & $0.067$ \\
      +Dec.. & $0.498$ & $29.45$ & $0.957$ & $0.053$ & $0.060$\\
      +Sem.  & $\textbf{0.503}$ & $\textbf{27.81}$ & $0.957$ & $\textbf{0.035}$ & $\textbf{0.054}$\\
       \hline
   \end{tabular}}

   \label{Quantitative ablation results of cross-reenactment on VoxCeleb.}
   }
    \end{center}
    \vspace{-0.4cm}
\end{table}

\begin{figure}[t] 
\centering 
    \includegraphics[width=0.9\linewidth]{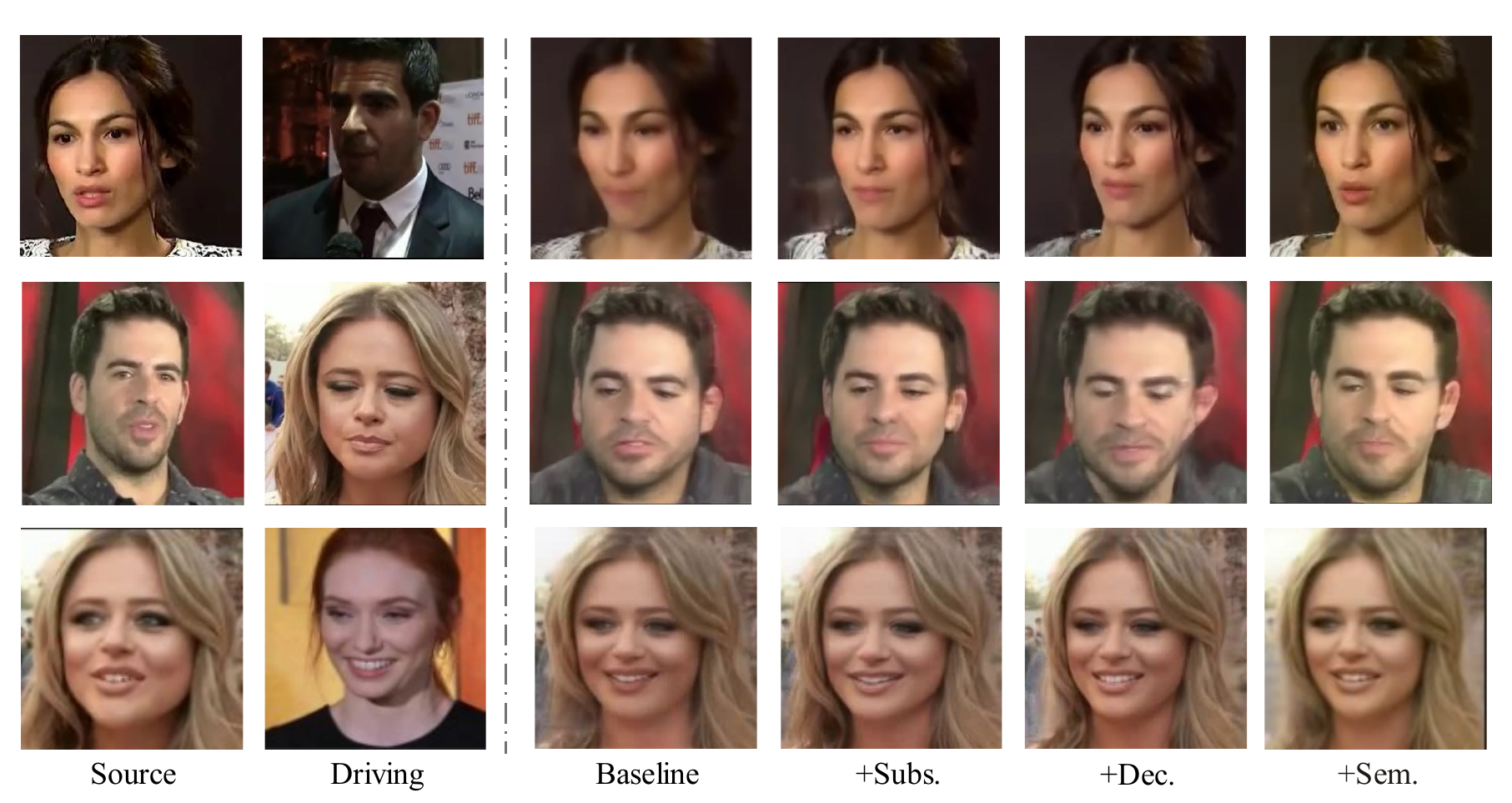}
    \vspace{-0.2cm}
    \caption{Qualitative results of the ablation study. We introduce the identity and motion subspaces (+Subs.), disentangle the two Subspaces (+Dec.) and enrich semantics (+Sem.) in both of them compared to the base model in turn.} 
    \vspace{-0.4cm}
    \label{ablation_study} 
\end{figure}

\section{Experiments}\label{sec:exps}
\subsection{Experimental Setup}\label{sec:exp_setup}

\begin{figure}[ht] 
    \centering 
    \includegraphics[scale=0.35]{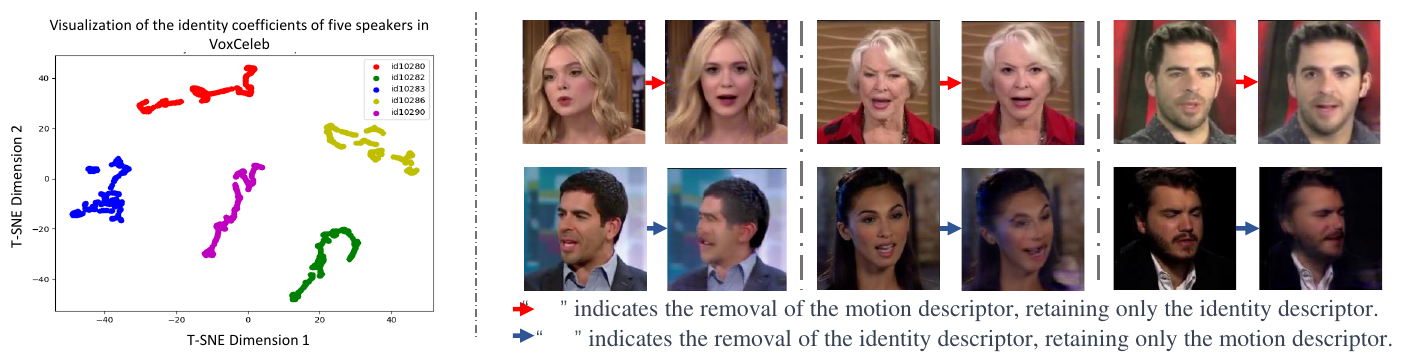} 
    \vspace{-0.3cm}
    \caption{Visualization for the Subspace Inseption Analysis.} 
    \vspace{-0.3cm}
    \label{id_visualization} 
\end{figure}

\begin{figure}[t] 
\centering 
\includegraphics[scale=0.26]{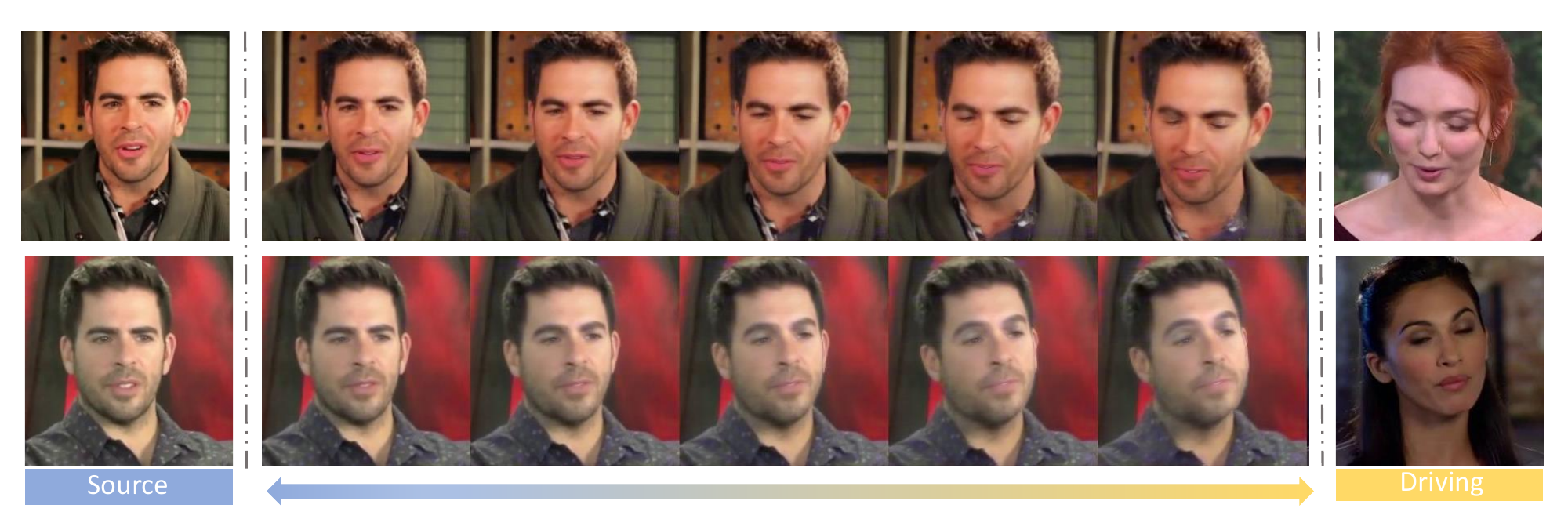} 
\vspace{-0.6cm}
\caption{Interpolation results between 2 speaking styles.}
\label{interpolation_case} 
\vspace{-0.6cm}
\end{figure}

\textbf{Dataset} We conducted experiments on three public datasets: VoxCeleb (256-resolution) \cite{VoxCeleb}, CelebV \cite{wu2018reenactgan}  (256-resolution) and HDTF \cite{HDTF}  (512-resolution). It is noteworthy that all our test sets and their corresponding training sets exhibit no overlap in terms of identities. 


\textbf{Evaluation Metrics}  For image quality, the Structural Similarity (\textbf{SSIM}) measures the low-level similarity of the generated images to the ground-truth images. The Fréchet Inception Distance (\textbf{FID}) \cite{FID} metric is employed to measure the dissimilarity between distributions of generated and real images. Of particular importance, we compute the cosine similarity of identity (\textbf{CSIM}) features to assess the fidelity of identity preservation. These features are derived from the pretrained face recognition model ArcFace \cite{ArcFace}. Furthermore, following the previous work PIRenderer \cite{PIRendererCP}, the Average Expression Distance (\textbf{AED}) and Average Pose Distance (\textbf{APD}) metrics are employed to scrutinize the impact of gesture and expression imitation.

\textbf{Implementation Details} 
The identity and motion subspaces each consist of 20 basis vectors, with 512 dimensions per basis vector.


\subsection{Quantitative Evaluation}
We conducted comparative evaluations between our method and several state-of-the-art (SOTA) approaches: FOMM \cite{FOMM}, PIRenderer \cite{PIRendererCP}, StyleHEAT \cite{StyleHEATOH}, DaGAN \cite{hong2022depth}, LIA \cite{LIA} and DPE \cite{pang2023dpe}.

The model trained on the VoxCeleb exhibits quantitative evaluation results for both the VoxCeleb and the CelebV test dataset, as depicted in Tab.\ref{Quantitative comparisons on the VoxCeleb dataset and CelebV dataset.}. Our method achieves the best performance among most metrics on VoxCeleb and CelebV on cross-identity facial animation. The significantly improved FID indicates that our image distribution is more similar to the target. The enhanced CSIM score suggests that our approach better preserves the identity of the source image. The exceptional AED and APD values signify that our method achieves more accurate motion transfer. We extended our model to a resolution of 512 and conducted training using the HDTF dataset at this resolution. We performed evaluations on both the same-identity and cross-identity facial animation tasks. The quantitative outcomes of our experimentation are detailed in Tab.\ref{Quantitative comparisons on the HDTF dataset}.

\subsection{Qualitative Evaluation}
We compare our method with several SOTA methods. The results are displayed in Fig.\ref{Qualitative comparisons with SOTA methods on Cross-reenactment.}. Our approach better preserves source identity and enables smoother motion transfer on VoxCeleb and HDTF dataset, resulting in highly realistic and natural-looking generated images. 
Although LIA exhibits favorable motion transfer results on low-resolution datasets, its ability to preserve identity remains insufficient. 
We hypothesize that this phenomenon arises from the fact that the “motion direction vectors" learned by LIA lack explicit semantic information and are coupled with features from the source image. This dual effect adversely affects both the preservation of the source image's identity and the precision of the learned motion vectors. While DPE yields promising results in expression transfer, the identity and motion are not completely decoupled.

\subsection{Ablation Study}
\noindent\textbf{Effectiveness of the various sub-methods.} We perform ablation experiments on the VoxCeleb dataset to investigate the effectiveness of the various sub-methods. 
We add the proposed sub-methods to the base model in turn and report the quantitative results in Tab.\ref{Quantitative ablation results of cross-reenactment on VoxCeleb.}. 
The improvement in CSIM, AED, and APD demonstrates that the introduction of identity and motion subspaces leads to a noticeable enhancement in preserving identity and facilitating accurate motion transfer. Upon separating the two subspaces, there is a further advancement in CSIM, as well as AED and APD. Strengthening the semantics in both subspaces enables a comprehensive expression of identity and motion, ultimately achieving high-fidelity identity preservation and precise motion migration. Throughout this process, the FID value consistently improves, indicating increased alignment distribution between the generated image and the target.

The qualitative results are shown in Fig.\ref{ablation_study}.
It can be seen that our base model suffers from severe facial distortion, challenges in maintaining identity, and inadequate motion transfer. The introduction of the two subspaces brings a more accurate face shape; however, inaccurate identity representation and imperfect motion emulation persist. By decoupling the two spaces, the generated image's identity and motion subspaces are isolated from irrelevant information. 


\subsection{Subspaces Inspection} 
\noindent\textbf{Identity Subspace Visualization}
We project the identity descriptors to a 2D space using t-SNE, which is commonly used in representation learning area~\cite{t-SNE,li2025decoupled,li2025freestyleret,li2023tg,li2022joint}. We select the identity descriptors of 5 different persons from the VoxCeleb test dataset. For each identity, we randomly select 200 frames to extract identity descriptors.
In the left of Fig.\ref{id_visualization}, the identity descriptors of the same person cluster in the identity subspace. This implies the identity descriptors of one person are similar. 
\noindent\textbf{Distinct Roles of the Two Descriptors}
To provide a more intuitive demonstration of the impact of identity and motion descriptors, we randomly selected two sets of photos from the test dataset. For the first set (upper row of the right in Fig.\ref{id_visualization}), we exclusively extracted their identity descriptors while setting the motion descriptors to zero, it is evident that the generated image preserves the original identity and maintains a relatively neutral expression and pose. For the second set (lower row of the right in Fig.\ref{id_visualization}), we extracted their motion descriptors while setting the identity descriptors to zero, it can be observed that the motion descriptor accurately retains motion, while an identity descriptor set to zero signifies that the image loses its facial shape and texture associated with identity.

\noindent\textbf{Motion Manipulation}
Thanks to the meaningful and linear motion subspace~\cite{jin2025local,jin2024hierarchical,lv2024navigating}, we can edit the motion by manipulating motion descriptors. As shown in Fig.\ref{interpolation_case}, when linearly interpolating between two motion descriptors extracted from different persons, the motion change of generated images transitions smoothly. 

\section{Conclusion}\label{sec:conclusion}

In this paper, We propose the semantic facial descriptors in learnable disentangled vector space to animate faces. In contrast to previous works, we successfully address the model-based methods’ limitations in high-fidelity identity, and the challenge faced by model-free methods in accurate motion transfer. Extensive results demonstrate that our model far outperforms SOTA methods with superior identity preserving and motion transfer.

\section{Acknowledgement}
The work is supported by National Science and Technology Major Project (2024ZD01NL00101) and Guangdong Provincial Key Laboratory of Ultra High Definition Immersive Media Technology(Grant No. 2024B1212010006). We thank all reviewers for their valuable comments.



\end{document}


\title{Dual-Process Watermarked Diffusion: Integrating Watermarking with Denoising in Point Cloud Models\\
-Supplementary Materials-
}


\author{Anonymous Authors}

\maketitle

\section{additional results}
In our paper, we demonstrate consistency, bit accuracy, and robustness to different attacks of point clouds generated from a watermarked diffusion model. 
The watermarked diffusion model is fine-tuned on a pre-trained watermark extractor, PI-HiDDeN. When we distill and fine-tune the diffusion model from the watermark extractor,
there will be some reduction in effectiveness inevitably. 
In practice, as an individual watermarking module for point clouds, the watermark extractor can be used directly to watermark point clouds with higher consistency and bit accuracy as a post-hoc watermarking method. In this section, we evaluate the performance of the PI-HiDDeN.

By watermarking the validation sets, we further evaluate the consistency (CD\&CMD) and bit accuracy of the watermarked point clouds under 8-bit payloads.
\autoref{tab: hidden evaluation} and \autoref{fig: hidden results} show the quantitative and qualitative results of PI-HiDDeN respectively. 
As a single watermark module, PI-HiDDeN shows better effectiveness than the watermarked diffusion model. Although the post-hoc watermarking method doesn't need to modify the network parameters of generative models, and the effect is better, it deviates from our original intention because the plagiarist who obtained the checkpoints of generative models could remove the post-hoc watermarking module manually.

\begin{table}[h]
\caption{Consistency and bit accuracy of PI-HiDDeN: We evaluate the CD and EMD between watermarked and original point clouds. It numerically shows that there is not too much difference between the two while ensuring high bit accuracy.}
\begin{tabular}{@{}llll@{}}
\toprule
Class    & Bit Acc(\%)$\uparrow$ & CD(\%)$\downarrow$  & EMD(\%)$\downarrow$ \\ \midrule
Airplane & 0.995       & 0.00329 & 0.550   \\
Table    & 0.995       & 0.00603 & 0.736   \\
Chair    & 0.998       & 0.00554 & 0.682   \\ \bottomrule
\end{tabular}
\label{tab: hidden evaluation}
\end{table}
\begin{figure}[t!]
    \begin{subfigure}[t]{0.5\linewidth}
        \centering
        \includegraphics[scale=0.5, page=1]{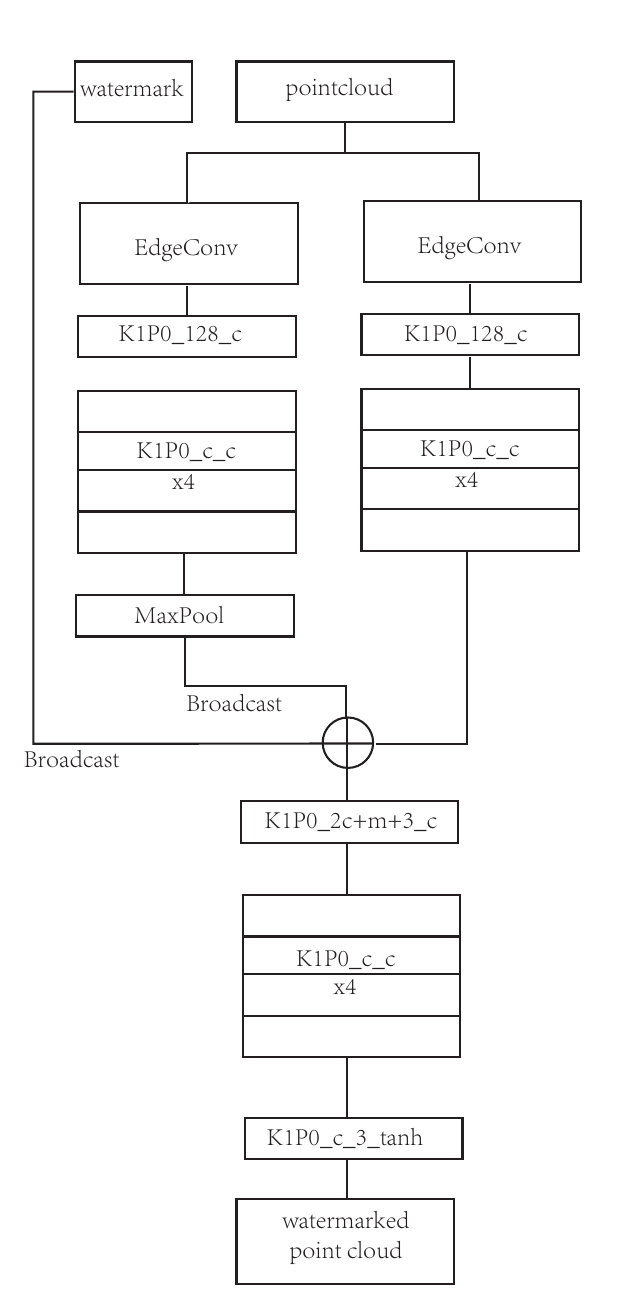}
        \caption{Watermark Encoder}
        \label{fig: encoder}
    \end{subfigure}%
    ~
    \begin{subfigure}[t]{0.5\linewidth}
        \centering
        \includegraphics[scale=0.5, page=2]{images/architecture.pdf}
        \caption{Watermark Decoder}
        \label{fig: decoder}
    \end{subfigure}
    \caption{The detailed architecture of PI-HiDDeN. }
    \label{fig: architecture}
\end{figure}

\section{architecture of Watermark Extractor}
The PI-HiDDeN consists of two parts: a watermark encoder and a watermark decoder as shown in \autoref{fig: architecture}. Given a \textit{m}-dimensional watermark and a point cloud with 3D coordinates, the watermark encoder will embed the watermark in the point cloud and then we can extract the watermark by the watermark decoder.

The convolution layer as K\{$k$\}P\{$p$\}\_\{$c_\text{in}$\}\_\{$c_\text{out}$\}\_\{$af$\}, where $k$ is the kernel size, $p$ is the padding size, $c_\text{in}$ and $c_\text{out}$ are the dimensions of the input and output channel respectively, and $ac$ is the activation function. Similarly, the L\_\{$c_\text{in}$\}\_\{$c_\text{out}$\}\_\{$af$\} represents the full connection layer. The $ac$ not given is by default GELU.
Lastly, $c$ in figures is a hyperparameter, determining the number of channels of intermediate data. In our experiments, $c$ is set to 128.

\section{Baseline Implementations}
In our experiments, we compare different backbones as our baseline such as PointNet, Transformer, and DGCNN. However, Some of the network framework parameters are relatively large, causing the training to fail to converge, so we made corresponding modifications to them.
For each backbone, we replace the EdgeConv module in \autoref{fig: architecture} for the watermark encoder and decoder synchronously.

\begin{figure*}[h]
  \centering
  \includegraphics[width=\textwidth]{images/hidden results.pdf}
  \caption{Visual results of watermarked point clouds by PI-HiDDeN. Most points of objects are consistent with the original points, and a small number of points move slightly to produce a watermark.}
   \label{fig: hidden results}
\end{figure*}

\paragraph{PointNet}
We modify the PointNet based on the implementation\footnote{Pointnet\_Pointnet2\_pytorch \url{https://github.com/yanx27/Pointnet_Pointnet2_pytorch}}, in which the \textit{PointNetEncoder} is used as our backbone. 
In practice, we changed some modules as follows:
\begin{itemize}
    \item Channels in \textit{PointNetEncoder}
    \begin{itemize}
        \item convolutions: [64, 128, 1024] $\rightarrow$ [128, 128, 128]
    \end{itemize}
        \item Channels in \textit{STN3d}
        \begin{itemize}
            \item convolutions: [64, 128, 1024] $\rightarrow$ [128, 128, 128]
            \item fc layers: [512, 256, 9] $\rightarrow$ [128, 128, 9]
        \end{itemize}
    \item \textit{global\_feat}=True
    \item \textit{feature\_transform}=False
\end{itemize}

\paragraph{Transformer}
We utilize the \textit{StackedAttention} in the implementation of Point Transformers \footnote{Point-Transformers \url{https://github.com/qq456cvb/Point-Transformers}}. There are two main changes in our experiments:
1) the number of channels we used is set to 128 and 2) only a single SA\_layer instead of its original quadruple ones is used.

\paragraph{DGCNN}
Based on the implementation of DGCNN for point clouds\footnote{DGCNN \url{https://github.com/WangYueFt/dgcnn}}, 
we use a single EdgeConv as our feature extractor, in which the number of nearest neighbor $k$ is set to 40 and the number of channels of the only convolution is set to 128.